\documentclass[letterpaper]{article} %DO NOT CHANGE THIS
\usepackage{aaai17}  %Required
\usepackage{times}  %Required
\usepackage{helvet}  %Required
\usepackage{courier}  %Required
\usepackage{url}  %Required
\usepackage{graphicx}  %Required
\usepackage[usenames,dvipsnames]{xcolor}
\graphicspath{{./figures/}}
\begin{document}
\newcommand{\ce}[1]{\texttt{\small{#1}}}

\title{Stakeholders in Explainable AI}
\author{Alun Preece \and Dan Harborne \\
Crime and Security Research Institute \\
Cardiff University, UK \\
Email: \{PreeceAD$|$HarborneD\}@cardiff.ac.uk
\And 
Dave Braines \and Richard Tomsett \\
IBM Research \\ 
Hursley, Hampshire, UK \\
Email: \{dave\_braines$|$rtomsett\}@uk.ibm.com
\AND 
Supriyo Chakraborty \\
IBM Research \\ 
Yorktown Heights, New York, USA \\
Email: supriyo@us.ibm.com}

\newcommand{\aedit}[1]{{\color{blue} #1}}%blue

\maketitle

\begin{abstract}
There is general consensus that it is important for artificial intelligence (AI) and machine learning systems to be explainable and/or interpretable. However, there is no general consensus over what is meant by `explainable' and `interpretable'. In this paper, we argue that this lack of consensus is due to there being several distinct stakeholder communities. We note that, while the concerns of the individual communities are broadly compatible, they are not identical, which gives rise to different intents and requirements for explainability/interpretability. We use the software engineering distinction between validation and verification, and the epistemological distinctions between knowns/unknowns, to tease apart the concerns of the stakeholder communities and highlight the areas where their foci overlap or diverge. It is not the purpose of the authors of this paper to `take sides' --- we count ourselves as members, to varying degrees, of multiple communities --- but rather to help disambiguate what stakeholders mean when they ask `Why?' of an AI.
\end{abstract}

\section{Introduction}

Explainability in artificial intelligence (AI) is not a new problem, nor was it ever considered a solved problem. The issue first came to prominence during the `knowledge engineering era' of the late 1970s and early 1980s, when the focus was on building expert systems to emulate human reasoning in specialist high-value domains such as medicine, engineering and geology~\cite{Buchanan:1984}. It was soon realised that explanations were necessary for two distinct reasons: system development, particularly testing, and engendering end-user trust~\cite{Jackson:1999}. Because the systems were based on symbolic knowledge representations, it was relatively straightforward to generate symbolic traces of their execution. However, these traces were often complex and hard for developers to interpret, while also being largely unintelligible to end-users because the reasoning mechanisms of the system were unrecognisable to human subject-matter experts. The latter problem led to approaches aimed at re-engineering knowledge bases to make the elements of the machine reasoning more recognisable, and to make the generated explanations more trustworthy~\cite{Swartout:1991}. These latter, `stronger' approaches to explainable AI were arrived at only when the knowledge engineering boom was effectively over, so they gained little traction at the time.

The last decade has seen a number of significant breakthroughs in machine learning via deep neural network approaches that has reinvigorated the AI field~\cite{LeCun:2015}. In this generation of AI development, the issue of explainability has again come into focus, though the term \textit{interpretability} is nowadays more commonly used, indicating an emphasis on humans being able to interpret machine-learned models. As in the 1970s and 1980s, there are differing motives between system developers and users in seeking explanations from an AI system: the former want to verify how the system is working (correctly or otherwise) while the latter want assurance that the outputs of the system can be trusted~\cite{Ribeiro:2016}. Unlike classical expert systems, deep neural network models are not symbolic so there is no prospect of generating intelligible `reasoning traces' at the level of  activation patterns of artificial neurons. Consequently, a distinction has been made between interpretability approaches that emphasise \textit{transparency} and those that are \textit{post-hoc}~\cite{Lipton:2016}. The former are explanations expressed in terms of the inner workings of a model while the latter are explanations derived `after the fact' from the workings of the model, such as an explanation in terms of similar `known' examples from the training data.

However, terminology in relation to explainability in modern AI is far from settled. A recent UK Government report on the state of AI received substantial expert evidence and noted, `The terminology used by our witnesses varied widely. Many used the term transparency, while others used interpretability or `explainability', sometimes interchangeably. For simplicity, we will use `intelligibility' to refer to the broader issue'~\cite{Lords:2017}. Others have used the term \textit{legibility}~\cite{Kirsch:2017} while recent thinking once again emphasises `strong' notions of explainability in causal terms~\cite{Pearl:2018}. Terminology is further complicated by concerns over the accountability~\cite{Diakopoulos:2016} and fairness~\cite{O'Neil:2016} of modern AI systems which, while overlapping the issue of end-user trust, extend into ethical and legal domains. These various perspectives and distinct groups of stakeholders have led to the rapid creation of a large and growing body of research, development, and commentary. Recent work seeks to place the field on a more rigorous scientific and engineering basis by, for example, examining axiomatic approaches to model interpretability~\cite{Leino:2018,Sundararajan:2017}, exploring more sophisticated methods for revealing the inner workings of deep networks~\cite{Olah:2018}, and arguing for increased use of theoretical verification techniques~\cite{Goodfellow:2018}. 

In summary, today there is a large community focused on the problem of explainable AI, with some seeking to advance the state of the art, others seeking to assess, critique, or control the technology, and still others seeking to exploit and/or use AI in a wide variety of applications. In our own recent work, we examined explainability and interpretability from the perspective of explanation recipients, of six kinds~\cite{Tomsett:2018}: system \textit{creators}, system \textit{operators}, \textit{executors} making a decision on the basis of system outputs, \textit{decision subjects} affected by an executor's decision, \textit{data subjects} whose personal data is used to train a system, and system \textit{examiners}, e.g., auditors or ombudsmen. We found this \textit{Interpretable to whom?} framework useful in thinking about what constitutes an acceptable explanation or interpretation for each type of recipient. In this paper, we take a slightly different tack, examining the stakeholder communities around explainable AI, and arguing that there are useful distinctions to be made between stakeholders' motivations, which lead to further refinement of the classical AI distinction between developers and end-users.

\section{Four Stakeholder Communities}

\subsubsection{Developers:} people concerned with building AI applications. Many members of this community are in industry --- large corporates and small/medium enterprises --- or the public sector, though some are academics or researchers creating systems for a variety of reasons including to assist them with their work. This community uses both terms `explainability' and `interpretability'. Their primary motive for seeking explainability/interpretability is quality assurance, i.e., to aid system testing, debugging, and evaluation, and to improve the robustness of their applications. They may use open source libraries created for generating explanations; some well-known and widely-used examples include LIME~\cite{Ribeiro:2016}, deep Taylor decomposition~\cite{Montavon:2016}, influence functions~\cite{Koh:2017} and Shapley Additive Explanations~\cite{Lundberg:2016}. Members of the developer community may have created their own explanation-generating code, motivated by an aim to aid practical system development rather than to advance AI theory. In terms of our \textit{Interpretable to whom?} framework, members of the developer community are system creators.

\subsubsection{Theorists:} people concerned with understanding and advancing AI theory, particularly around deep neural networks. Members of this community tend to be in academic or industrial research units. Many are also active practitioners, though the theorist community is distinguished from developers by their chief motivation being to advance the state of the art in AI rather than deliver practical applications. Members of the theorist community tend to use the term `interpretability' rather than `explainability'. The motive to better understand fundamental properties of deep neural networks has led to some interpretability research being labelled `artificial neuroscience'~\cite{Voosen:2017}. A well-known early piece of work identified properties of activation patterns, and also how deep neural networks are vulnerable to adversarial attacks~\cite{Szegedy:2014}. Recent work in this milieu has looked at feature visualisation to better interpret properties of hidden layers in deep networks~\cite{Olah:2018}. It has also been suggested that such interpretations may provide new kinds of cognitive assistance to human understanding of complex problem spaces~\cite{Carter:2017}. Membership of this community of course overlaps with the developer community. For example, in the case of an industry researcher who carries out theoretical work on deep neural network technology (theorist) while also applying the technology to build systems (developer). In our `Interpretable to whom?' framework, members of the theorist community are considered system creators.
 
\subsubsection{Ethicists:} people concerned with fairness, accountability and transparency\footnote{`Transparency' in the common usage of the term rather than the specific usage by~\cite{Lipton:2016} and others.} of AI systems, including policy-makers, commentators, and critics. While this community includes many computer scientists and engineers, it is widely interdisciplinary, including
social scientists, lawyers, journalists, economists, and politicians. As well as using `explainability' and `interpretability', members of this community use `intelligibility' and `legibility' as noted in the introduction. A subset of this community will also be members of the developer and/or theorist communities\footnote{Indeed, professional bodies including ACM, BCS and IEEE all place significant emphasis on recognising ethical, legal and societal issues in software development.} but their motives in seeking explanations are different: for the ethicist community, explanations need to go beyond technical software quality to provide assurances of fairness, unbiased behaviour, and intelligible transparency for purposes including accountability and auditability --- including legal compliance in cases such as the European Union's GDPR legislation~\cite{Goodman:2016}. Our \textit{Interpretable to whom?} framework considers members of ethicist community to be dispersed across all six roles, though the distinct explanation-seeking motive of the ethicist community aligns most closely with system examiners, creators, data subjects and decision subjects.

\subsubsection{Users:} people who use AI systems. The first three communities comprise the vast majority of people who contribute to the growing literature on AI explainability/interpretability, whereas our fourth generally does not. Members of the user community need explanations to help them decide whether/how to act given the outputs of the system, and/or to help justify those actions. This community includes both `hands on' end-users but also everyone involved in processes that are impacted by an AI system. Consider an insurance company that uses an AI tool to help decide whether and at what cost to sell policies to clients. The end-users of the tool, the director of the company, and the clients are all members of the user community. Again, members of the user community may also be in other stakeholder communities, sometimes in relation to the same AI system; for example, an academic criminologist who has learned how to apply AI technology to create a predictive analytics tool (developer) to assist them in their research (user), while being aware of societal impacts of their work (ethicist). The \textit{Interpretable to whom?} framework places system operators and decision executors in the user community, along with decision subjects.\footnote{Arguably, decision subjects will be aligned with the user or ethicist communities, depending on how `empowered' they perceive themselves to be in relation to the effects of the system outputs.}

\section{Engineering and Epistemological Perspectives}

Explanation is closely linked to evaluation of AI systems. As noted in the introduction, early AI explanation efforts aimed to help system developers diagnose incorrect reasoning paths. Modern transparent interpretation methods are akin to such `traces', while post-hoc explanation techniques can be regarded as `diagnostic messages'. Moreover, explanations speak to issues of user trust and system impact, to the user and ethicist communities. Colloquially, in software engineering, \textit{verification} is about `building the system right' whereas \textit{validation} is about `building the right system'. In terms of explanation, verification is mainly associated with transparent techniques; `glass box' approaches are essential because it matters greatly how the system is built. Validation is more concerned with what the system does (and does not do) and so post-hoc techniques are often useful here. 

In line with this thinking, and at risk of overgeneralising, we assert that the developer and theorist communities tend to focus more on verification: the former because they want a system that is `built right', and the latter because they are interested understanding how the various kinds of deep neural networks work, and what are their theoretical limits. We suggest that the user and ethicist communities are more focused on validation, being more concerned with what an AI system does than about how it is built. This means that the developer and theorist communities tend to focus on transparency-based explanation techniques, while user and ethicist communities value post-hoc techniques.

From an epistemological perspective, we can consider the familiar framing in terms of knowns and unknowns: 

\textbf{Known knowns}: for an AI system based on machine learning, these constitute the set of available training and test data. The ability of the system to deal with known knowns is verified by standard testing approaches (e.g., $n$-fold cross-validation) and reported in terms of accuracy measures. Within the bounds of the known knowns, transparency-based explanation techniques such as deep Taylor decomposition~\cite{Montavon:2016} or feature visualisation~\cite{Olah:2018} can be used to `trace' the relationships between features (in input and hidden layers) and outputs. All four stakeholder communities have a clear interest in understanding the space of known knowns, though we would argue that it tends to be the developer constituency that are most focused on this space: maximising system performance within the space, defining the bounds of the space, and widening those bounds as much as is feasible.

\textbf{Known unknowns}: these constitute the space of queries, predictions, or behaviours that the AI system is intended to perform. The accuracy measures produced in system testing (verification) provide an estimate of the ability of the system to deal well with the space of known unknowns. The value of a system to members of the user community is in terms of this ability (otherwise the system is nothing more than a retrieval tool for known knowns). Feedback processes are needed because system system outputs may prove to be invalid at run-time (e.g., the system recommends an action that turns out to be inappropriate) leading to the generation of additional data for the training (known knowns) space. Members of the theorist community are interested in better understanding how AI systems process known unknowns~\cite{Olah:2018,Szegedy:2014}, and creating improved architectures for doing so.

\textbf{Unknown knowns}: from the perspective of the AI system, these are things that are outside its scope, but known more widely. Some biases of concern to the ethicist constituency fall into this category: a narrowness or skew in the training data results in a model that is `blind' to particular prejudices~\cite{Diakopoulos:2016,O'Neil:2016}. Validation is key to revealing such unknown knowns. 

\textbf{Unknown unknowns}: these have recently been highlighted as a key concern in AI system robustness~\cite{Dietterich:2017}, with a variety of methods being proposed to deal with them, including employing a portfolio of models to mitigate against weaknesses in individual models, and creating AI systems that build causal models of the world~\cite{Lake:2017} and/or or are aware of their own uncertainty~\cite{Kaplan:2018}. Clearly, all four communities have reason to be concerned with unknown unknowns: developers in terms of system robustness, theorists in terms of seeking stronger theories and architectures, ethicists in terms of ethical and legal implications of AI system failings, and users in terms of impacts on themselves and their livelihoods. 

In software engineering, formal verification techniques have been used to mathematically define the space of knowns --- in terms of a system specification --- leaving only the unknown unknowns fully excluded from that space. The theorist community is beginning to think along these lines~\cite{Goodfellow:2018}, though how to formally specify the intended behaviour of a deep neural network-based AI system remains an open question. This difficulty has been highlighted in recent years by research into `adversarial examples', which are designed to fool machine learning models by minimally perturbing input data to cause incorrect classifications~\cite{Goodfellow:2014,Szegedy:2014}. Such examples take advantage of the difficulty in learning correct classification decision boundaries from limited, high-dimensional data. While several methods to mitigate against such attacks have been proposed~\cite{Papernot:2015,Ross:2017a}, none amounts to a formal verification of the model's behaviour on adversarial inputs (though see \cite{Dvijotham:2018}). Building uncertainty awareness into models so that they can recognise and explicitly deal with such unknown unknowns may be a reliable way of improving system robustness~\cite{Gal:2018}, though with unkown implications for human interpretability.

\section{Explanation Types and Discussion}

\subsubsection{Transparency-based explanations:} The definition of transparency in~\cite{Lipton:2016} appears consistent with the notion of `full technical transparency' in~\cite{Lords:2017}. Both sources conclude that achieving full transparency is not realistic for anything other than small models, e.g., shallow decision trees or rule bases. A more limited form of transparency is exhibited by attribution techniques that visualise activations in the input or hidden layers of a network (e.g., deep Taylor decomposition~\cite{Montavon:2017}, feature visualisation~\cite{Olah:2018}) often as a \textit{saliency map} showing the features of the input that had most significance in determining the output. While noting that the visualisation element of these approaches is a post-hoc technique~\cite{Lipton:2016}, we nevertheless consider these methods \textit{transparency-based}, to distinguish them from `purely post-hoc' approaches that do not derive at all from inner states of the model.

\begin{figure}[t]
\centering
\includegraphics[width=0.48\textwidth]{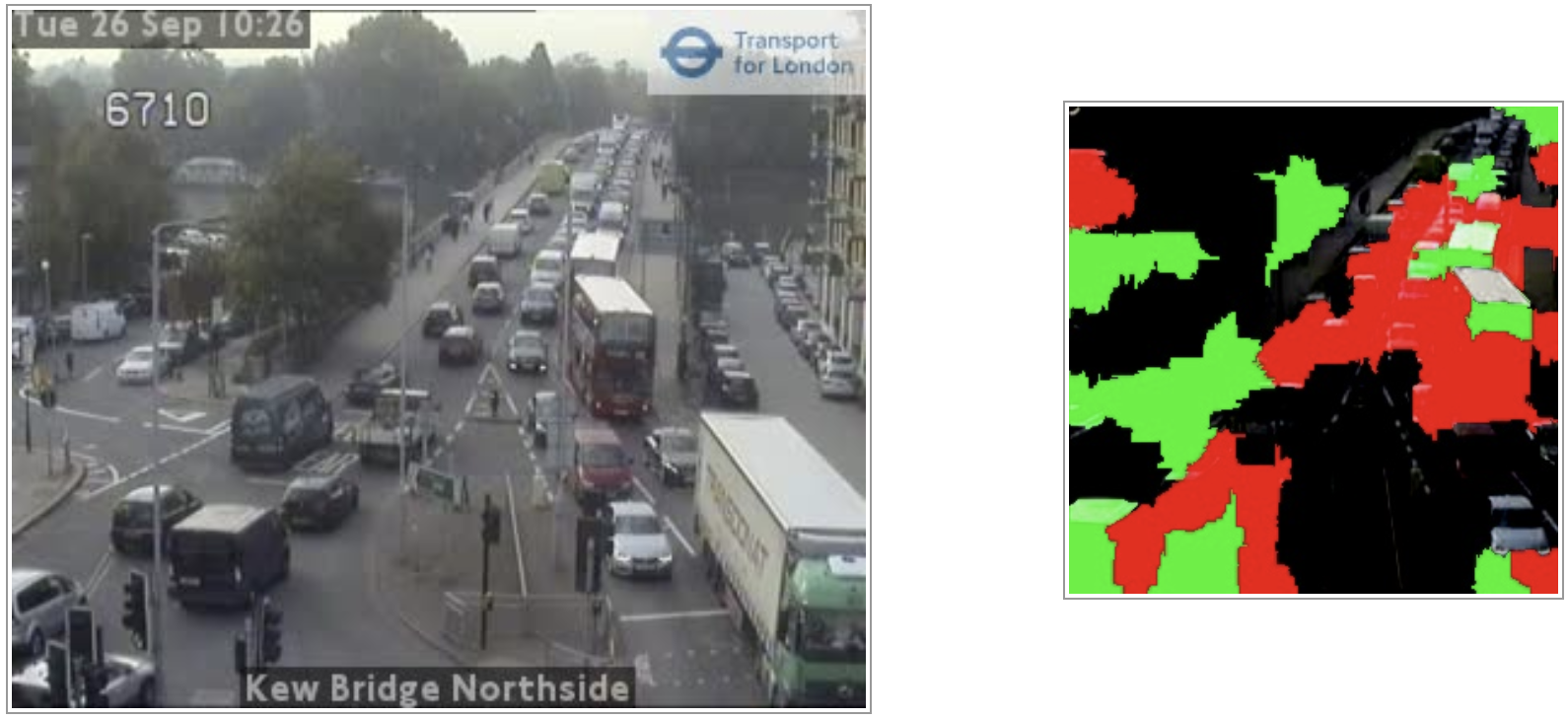}
\caption{Example saliency map for traffic congestion: the red regions of the input image are most significant in classifying the image as congested}
\label{fig:saliency}
\end{figure}

Figure~\ref{fig:saliency} shows an example saliency map for a traffic congestion monitoring system, from~\cite{Harborne:2018}\footnote{The example map was generated using the LIME software~\cite{Ribeiro:2016} which does not conform to our definition as being \textit{transparency-based} because it generates a local approximation of the learned model; it is included here only as an example of what a saliency map looks like in general.}. From a system verification perspective, such explanations would seem of immediate value to the developer and theorist communities, though with the caveat that many attribution methods are unstable~\cite{Sundararajan:2017} and/or unreliable~\cite{Kindermans:2017}. In addition to these technical concerns, attribution visualisations can be hard to interpret by members of the user and ethicist community where the explanation does not clearly highlight  meaningful features of the input. Therefore, such explanations are in danger of making members of these communities \textit{less} inclined to trust the system because they appear to reveal a system that operates in an unintelligible, unstable, `inscrutable' or `alien' manner. Even when an explanation seems `convincing' because it highlights meaningful and plausible features, there is a danger of confirmation bias in the receiver unless counterfactual cases are also included. Providing detailed transparency-based explanations may also overwhelm the recipient --- more information is not necessarily better for user performance~\cite{Marusich:2017}. 

\begin{figure}[t]
\centering
\includegraphics[width=0.48\textwidth]{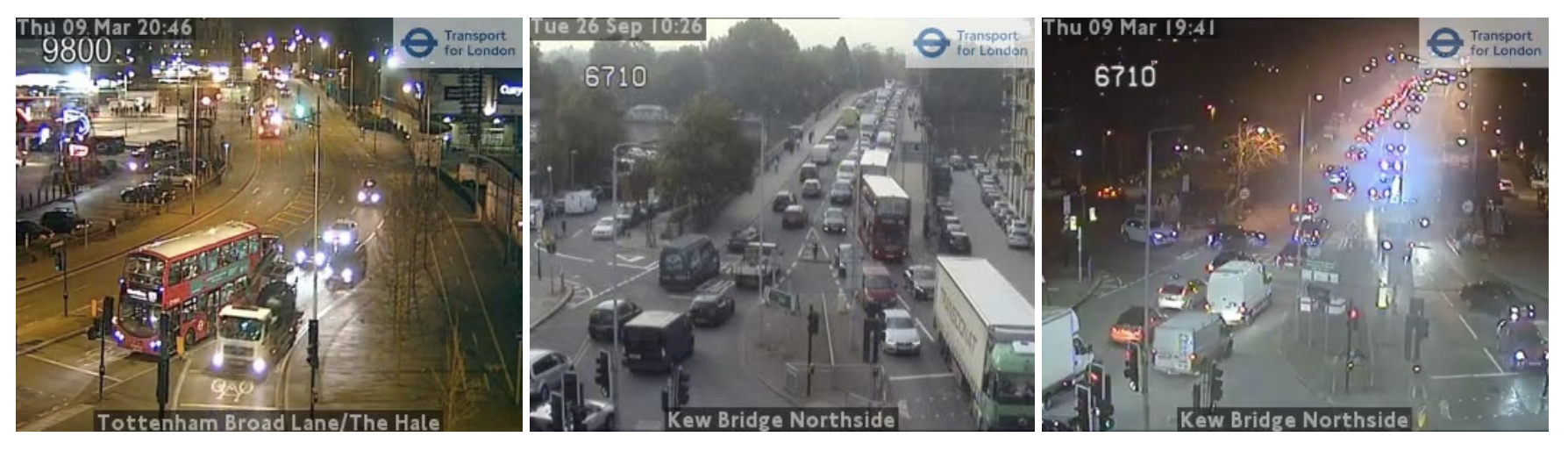}
\caption{Example explanation-by-example for a traffic congestion classification: the input image is in the middle; the left and right images are training examples with congestional classification probabilities slightly lower and higher, respectively, than the input}
\label{fig:ebe}
\end{figure}

\subsubsection{Post-hoc explanations:} A commonly-used type of post-hoc explanation is approximation using a local model, e.g., visualised as a saliency map as in LIME~\cite{Ribeiro:2016}, or in the form of a decision tree~\cite{Craven:1996}. Such techniques provide explanations that appear similar to those generated by transparency-based techniques and, if offered to users or ethicists, it is important to communicate clearly that they are actually post-hoc approximations. Explanations in terms of examples --- see Figute~\ref{fig:ebe} --- are a traditional approach favoured by subject-matter experts~\cite{Lipton:2016} and therefore especially appropriate for the user and ethicist communities. Approaches here include identifying instances from the training set most significant to a particular output~\cite{Koh:2017} and employing case-based reasoning techniques to retrieve similar training examples~\cite{Caruana:1999}. Such approaches have an advantage that counterfactual examples can also be provided. Another common post-hoc technique targeted towards users is to generate text explanations; the approach in~\cite{Hendricks:2016} uses background domain knowledge to train the system to generate explanations that emphasise semantically-significant features of the input.

\subsubsection{Layered explanations:} From the above discussion, it may seem that the sensible approach is to offer different explanations tailored to the different stakeholders, but can we envisage instead a composite \textit{explanation object} that packs together all the information needed to satisfy multiple stakeholders, and can be unpacked (e.g., by accessor methods) per a recipient's particular requirements. Moreover, we can view such an object being layered as follows:\\
\textbf{Layer 1 --- traceability}: transparency-based bindings to internal states of the model so the explanation isn't entirely a post-hoc rationalisation and shows that the system `did the thing right' [main stakeholders: developers and theorists]; \\
\textbf{Layer 2 --- justification}: post-hoc representations (potentially of multiple modalities) linked to layer~1, offering semantic relationships between input and output features to show that the system `did the right thing' [main stakeholders: developers and users]; \\
\textbf{Layer 3 --- assurance}: post-hoc representations (again, potentially of multiple modalities) linked to layer~2, with explicit reference to policy/ontology elements required to give recipients confidence that the system `does the right thing' (in more global terms than Layer~2) [main stakeholders: users and ethicists]. \\
\textbf{Example --- wildlife monitoring system}: Layer~1 (traceability): saliency map visualisation of input layer features for classification `gorilla'; Layer 2 (justification): `right for the right reasons' semantic annotation of salient gorilla features; Layer 3 (assurance): counterfactual examples showing that images of humans are not miss-classified as `gorilla'.

\section{Conclusion}

In this paper we have attempted to `tease apart' some of the issues in explainable AI by focusing on the various stakeholder communities and arguing that their motives and requirements for explainable AI are not the same. We related notions of transparent and post-hoc explanations to software verification and validation, and consideration of knowns/unknowns. We suggested that a `layered' approach to explanations that incorporates transparency with local and global post-hoc representations may serve the needs of multiple stakeholders. 

On a final note, the most influential of our four stakeholder communities is the users --- the one that's barely represented in the literature --- because, as in the 1980s, failure to satisfy users of AI technology in the long run will be the most likely cause of another `AI Winter'. Unfulfilled expectations and/or a smaller-than-hoped-for market will lead to investment drying up.

\end{document}